\documentclass[11pt]{article}
\usepackage[preprint]{acl}

\usepackage[utf8]{inputenc}
\usepackage[T1]{fontenc}
\usepackage{times}
\usepackage{latexsym}
\usepackage{url}
\usepackage{booktabs}
\usepackage{amsfonts}
\usepackage{nicefrac}
\usepackage{microtype}
\usepackage{xcolor}
\usepackage{xspace}
\usepackage{graphicx}
\usepackage{amsmath}
\usepackage{adjustbox}
\usepackage{tabularx}
\usepackage{multirow}
\usepackage{array}
\usepackage{makecell}
\usepackage{siunitx}
\usepackage{enumitem}
\usepackage{fancyvrb}
\fvset{fontsize=\scriptsize}
\usepackage{tikz}
\usetikzlibrary{arrows.meta,positioning}

\newcommand{\RENDER}{\textnormal{RENDER}\xspace}

\newcommand{\LongMemEval}{LongMemEval\xspace}
\newcommand{\LoCoMo}{LoCoMo\xspace}
\newcommand{\HotpotQA}{HotpotQA\xspace}

\newcommand{\LangChain}{LangChain\xspace}
\newcommand{\MemGPT}{MemGPT\xspace}
\newcommand{\ChatGPT}{ChatGPT\xspace}

\newcommand{\OpenAI}{OpenAI\xspace}
\newcommand{\Anthropic}{Anthropic\xspace}
\newcommand{\Google}{Google\xspace}
\title{RENDER: Controlling Reader-Facing Evidence in LLM Memory Evaluation}

\author{
Yuan Si \\
University of Waterloo \\
Waterloo, Canada \\
\texttt{yuan.si@uwaterloo.ca}
\And
Simeng Han \\
Stanford University \\
Stanford, USA \\
\texttt{shan6@law.stanford.edu}
\AND
Daming Li \\
Independent Researcher \\
USA \\
\texttt{damingliyale22@gmail.com}
\And
Jialu Zhang\thanks{Corresponding author}\\
University of Waterloo \\
Waterloo, Canada\\
\texttt{jialu.zhang@uwaterloo.ca}
}
\begin{document}

\maketitle

\begin{abstract}
Memory and RAG evaluations often treat the answering model's input as an implementation detail, even though systems may render the same history as a memory entry, summary, typed record, or raw excerpt.

We introduce \RENDER, a benchmark control that fixes the conversation while varying the reader-facing artifact. \RENDER combines a five-level packet ladder, localizing when answer-bearing content enters the input, with deterministic templates approximating \ChatGPT{}-style entries, \LangChain{} summaries, \MemGPT{}-style typed records, and raw conversation. On 500 \LongMemEval{} questions and nine models, matched-budget resolved packets beat recency-truncated raw dialogue by 42.4--72.6 points. In deployed-style templates, best--worst spread is 24.6--48.8 points per model; under the primary scorer, \ChatGPT{}-style entries have higher point estimates than raw conversation on 7 of 9 models. Judge rescoring preserves the positive aggregate effect, but model-specific significance is mixed. Three models scoring 0\% on formal ledger packets answer the same facts from natural-language entries at 45.4--53.4\%. The effect persists under retrieval noise and transfers to \HotpotQA{}, suggesting that memory/RAG evaluations should report or control the reader-facing artifact.

\end{abstract}

\section{Introduction}
\label{sec:intro}

A memory or retrieval-augmented generation (RAG) system does not hand the reader a conversation; it hands the reader a \emph{rendering} of one. We use rendering broadly: the reader-facing artifact includes selection, compactness, conflict-resolution exposure, metadata density, abstention cues, and surface form. Across diagnostic and matched-budget controls, we show that this choice is not a neutral implementation detail: measured accuracy ranges from near zero to 82\%, depending on what artifact reaches the answering model.

This is an evaluation problem. Memory and RAG benchmarks often compare pipelines whose readers receive different artifacts: one system passes raw retrieved dialogue, another passes a summary, and another passes a structured memory record. If those artifacts are not reported or controlled, the score conflates model ability, retrieval quality, memory-update logic, compactness, conflict-resolution exposure, abstention cues, and evidence rendering. A reported gain may therefore reflect a better reader-facing artifact rather than a better model or memory system. We ask: \emph{holding the underlying conversation, question, and answer contract fixed, how much can the reader-facing evidence artifact change measured memory-QA accuracy?}

Consider a user who first says they live in Boston, later says they moved to Denver, and then asks ``Where do I live now?'' A \ChatGPT{}-style memory entry may show \emph{``User lives in Denver (moved from Boston)''}; a \MemGPT{}-style record may expose \texttt{user.city = "Denver"}~\citep{memgpt2023}; a \LangChain{}-style summary may compress the session into a paragraph~\citep{langchainSummaryMemory2026}. These artifacts encode the same fact, but they are not equivalent reader inputs. In our experiments, deployed-style templates create a 24--49 point accuracy spread per model.

We introduce \RENDER (\emph{Reader Evidence Rendering Diagnostics}), a benchmark control that fixes the underlying conversation and answer contract while varying the reader-facing artifact. \RENDER has two instruments. First, a five-level packet ladder reveals answer-bearing content progressively: $P_0$ and $P_1$ expose only witness addresses, $P_2$ first writes the resolved current-state value into the packet body, and $P_3$/$P_4$ add metadata around that value. This ladder localizes whether failures stem from missing content, unresolved conflict, or the surrounding packet surface. Second, a template-family comparison renders the same underlying dialogue as deterministic approximations of deployed memory surfaces: \ChatGPT{}-style natural-language entries, \LangChain{} summaries, \MemGPT{}-style typed records, and raw conversation.

\RENDER is an input-side study, not an output-format study. Prior work asks whether models reason worse when required to \emph{produce} JSON, XML, or constrained outputs~\citep{formatrestrictions2024,lee2026formattax}. We fix the answer contract and vary only the evidence the reader sees.

We run about 238{,}000 model calls across \LongMemEval{} (500 questions; \citealt{longmemeval2024}), \LoCoMo{} (200 questions; \citealt{locomo2024}), retrieval-noise variants, and \HotpotQA{} transfer. The main experiments use nine commercial models from \OpenAI{}, \Anthropic{}, and \Google{}, with tools, search, and provider-native memory disabled. An independent LLM judge rescores the 31{,}500 items from the two headline experiments, with a 600-item second-judge sample confirming agreement (Appendix~\ref{app:judge}).

\paragraph{Findings.}
\begin{enumerate}[leftmargin=*,topsep=2pt,itemsep=1pt]
    \item \textbf{Accuracy rises when the answer enters the packet.} All nine models stay near 0\% at $P_0$/$P_1$; responsive models recover to 15--25\% at $P_2$. Metadata above $P_2$ changes accuracy by at most $\pm$2 points.
    \item \textbf{Budget control reverses the raw-vs-structured conclusion.} Full raw conversation outperforms budgeted formal packets, but under matched word budgets, streamlined resolved-$P_2$ packets outperform recency-truncated raw dialogue by 42.4--72.6 points on every model.
    \item \textbf{Deployed-style surface choice substantially changes the score.} Under the primary substring scorer, \ChatGPT{}-style entries have higher point estimates than raw conversation on 7 of 9 models; judge rescoring preserves a positive aggregate effect but shows mixed model-specific significance. \LangChain{} summaries score 4.4--10.6\%; \MemGPT{}-style typed records score 14.8--33.0\%. The best--worst spread reaches 24.6--48.8 points per model.
    \item \textbf{Evidence artifacts can induce refusal.} Three models score 0\% on formal ledger packets but answer the same facts from natural-language entries at 45.4--53.4\%, consistent with artifact/prompt-triggered abstention rather than missing capability.
\end{enumerate}

These results support a reporting norm: memory/RAG evaluations should state the reader-facing evidence artifact or include a fixed-artifact control. In our experiments, without this control, a 24--49 point artifact spread is indistinguishable from model or pipeline progress.

\section{Related Work}
\label{sec:related}

\paragraph{Long-context and memory evaluation.}
Long-context benchmarks stress retrieval, context length, positional robustness, or long-horizon reasoning rather than the rendered form of fixed evidence. Examples include RULER~\citep{ruler2024}, InfiniteBench~\citep{infinitebench2024}, L-Eval~\citep{leval2024}, ZeroSCROLLS~\citep{zeroscrolls2023}, LongBench and LongBench v2~\citep{longbench2024,longbenchv2_2024}, Marathon~\citep{marathon2024}, and BABILong~\citep{babilong2024}. Architecture-level work extends usable context through recurrence, compression, retrieval, or memorization~\citep{dai2019transformerxl,rae2020compressive,wu2022memorizing,khandelwal2020knn}. \LongMemEval{}~\citep{longmemeval2024} and \LoCoMo{}~\citep{locomo2024} are closest to our setting because they evaluate multi-session conversational memory with updates and temporal scope. Work on position sensitivity shows that models may fail even when relevant information is present~\citep{lostinmiddle2024}; \RENDER studies an orthogonal variable: how that information is rendered to the reader.

\paragraph{Memory, retrieval, and RAG systems.}
Modern memory systems often rewrite raw histories before generation. \MemGPT{}~\citep{memgpt2023} externalizes memory into an OS-like hierarchy; HippoRAG and GraphRAG organize evidence through graph indices~\citep{hipporag2024,graphrag2024}; ENGRAM uses typed memory schemas~\citep{engram2025}; and agent memory work includes MemoryBank, Generative Agents, and Reflexion~\citep{memorybank2023,park2023generativeagents,reflexion2023}. Production libraries such as \LangChain{} also expose summary-based conversation memory utilities~\citep{langchainSummaryMemory2026}. RAG work attaches an external store to the reader~\citep{rag2020,realm2020,retro2022,atlas2022,Gmerge}, extends retrieval to multi-hop reasoning~\citep{ircot2023,selfrag2024}, and evaluates retrieval quality, faithfulness, and context use~\citep{ragas2024}. These studies generally evaluate end-to-end systems; \RENDER isolates the reader-facing artifact itself.

\paragraph{Format sensitivity and judging.}
Output-side structured-generation work changes how the model must answer, e.g., by requiring JSON, XML, Markdown, or \LaTeX{}~\citep{formatrestrictions2024,lee2026formattax}. \RENDER changes what evidence the model reads while holding the answer contract fixed. More broadly, LLMs are sensitive to prompt formatting, few-shot ordering, demonstration composition, chain-of-thought prompts, and self-consistency decoding~\citep{sclar2024prompt,lu2022fantastically,min2022rethinking,zhao2021calibrate,cot2022,kojima2022,selfconsistency2023,Pydex}; recent RAG work also shows that context formatting and stronger raw-passage baselines can change downstream reasoning and evaluation conclusions~\citep{contextualnormalization2025,strongerbaselinesrag2025}. Tooling for constrained generation helps systems emit typed records~\citep{outlines2023}, but does not test whether those records are the best evidence artifact for a downstream reader. We use substring matching, normalized exact match, and LLM judges, building on work on MT-Bench/Chatbot Arena, G-Eval, PandaLM, and known judge biases~\citep{mtbench2023,geval2023,pandalm2024,fairevaluators2024,NETTLE,gu2026context}. \HotpotQA{} and related multi-hop QA datasets provide a non-conversational transfer setting~\citep{yang2018hotpotqa,wikimultihop2020,musique2022}. Abstention and knowledge-conflict work shows that refusal behavior can be shaped by task framing and contextual conflict~\citep{rtuning2024,falsereject2025,abstentionsurvey2025,xstest2024,kadavath2022know,tian2023calibration,lin2022teaching,knowledgeconflicts2024,beliefrevision2024,longpre2021entity,xie2024adaptive}; our contribution is to show that reader-facing artifacts and their abstention cues can trigger such behavior.

\section{System Design and Methodology}
\label{sec:method}

\RENDER is a benchmark control for a narrow input-side question: when the underlying conversation, question, and answer contract are fixed, how much can the reader-facing evidence artifact change memory-QA accuracy? Each item has three fixed objects: the original dialogue, the question, and the expected answer behavior. \RENDER varies the artifact shown to the final answering model, which we call the \emph{reader}; where a control also changes prompt wrapping, we label that as part of the condition. The reader receives one rendered evidence artifact, answers the same free-form question, and is forbidden from using external tools, search, provider-native memory, or hidden context.

Internally, \RENDER stores turn-addressed dialogue, identifies answer-bearing events, resolves conflicts, and renders different reader-facing artifacts. Externally, the reader sees only the rendered artifact and a fixed free-form answer contract. No condition asks the reader to emit JSON, XML, typed records, or any other structured output; the manipulated variable is the input-side evidence artifact.

\paragraph{Packet ladder.}
The first control is a five-level ladder that reveals increasingly more answer-bearing information from the same evidence. The ladder is a processing-stage diagnostic: it identifies whether failure occurs because the reader cannot see relevant content, because conflict resolution has not been made explicit, or because the surrounding packet surface changes reader behavior. We use: $P_0$ \texttt{raw\_packet}, with witness addresses and blocker flags but no answer text; $P_1$ \texttt{positive\_only}, with witness addresses only; $P_2$ \texttt{active\_only}, the first level that writes the resolved current-state value into the packet body; $P_3$ \texttt{ctx\_normalized}, which adds blocker metadata around $P_2$; and $P_4$ \texttt{full\_render}, which adds a budget-aware list of supporting entries. $P_0$/$P_1$ expose audit pointers, not links the reader can open. $P_2$ is therefore the first answer-bearing packet.

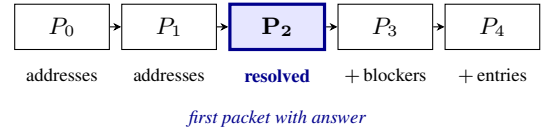
\begin{figure}[t]
\centering
\begin{tikzpicture}[
    box/.style={rectangle, draw, minimum height=0.6cm, minimum width=1.25cm, align=center, font=\footnotesize},
    boxhi/.style={box, fill=blue!10, very thick, draw=blue!55!black},
    arr/.style={-{Stealth[length=3pt,width=3pt]}, thin},
    desc/.style={font=\scriptsize, align=center},
]
\node[box] (p0) {$P_0$};
\node[box, right=0.15cm of p0] (p1) {$P_1$};
\node[boxhi, right=0.15cm of p1] (p2) {$\mathbf{P_2}$};
\node[box, right=0.15cm of p2] (p3) {$P_3$};
\node[box, right=0.15cm of p3] (p4) {$P_4$};
\draw[arr] (p0) -- (p1);
\draw[arr] (p1) -- (p2);
\draw[arr] (p2) -- (p3);
\draw[arr] (p3) -- (p4);
\node[desc, below=3pt of p0] {addresses};
\node[desc, below=3pt of p1] {addresses};
\node[desc, below=3pt of p2, text=blue!55!black, font=\scriptsize\bfseries] (d2) {resolved};
\node[desc, below=3pt of p3] {$+$\,blockers};
\node[desc, below=3pt of p4] {$+$\,entries};
\node[below=3pt of d2, font=\scriptsize\itshape, text=blue!55!black] {first packet with answer};
\end{tikzpicture}
\caption{The five-packet ladder. $P_0$/$P_1$ expose witness addresses only; $P_2$ first writes the resolved current-state answer into the packet body; $P_3$/$P_4$ add blocker metadata and supporting entries.}
\label{fig:packet_ladder}
\end{figure}

\paragraph{Queries and conflict resolution.}
Each dialogue is ingested into an append-only store with stable turn addresses of the form \texttt{history/<session>/turn/<turn>}. Each question is mapped to a canonical \texttt{RenderQuery} containing a normalized slot key, temporal scope (e.g., current-state or as-of reasoning), and conversational scope (e.g., full-history or session-restricted reasoning). Candidate events are tagged as updates, corrections, contradictions, or deletions, then ordered by precedence; deletions take highest precedence. Scope resolution marks candidates as \emph{applicable}, \emph{out\_of\_scope}, or \emph{blocked}. A resolved current state exists only when the slot collapses to one applicable non-blocked event. Otherwise the packet carries a blocker, either \texttt{scope\_blocked} or \texttt{unresolved\_conflict}. Abstention is rendered explicitly as
\[
\texttt{should\_abstain} = \mathbf{1}[\#\text{blockers} > 0].
\]

\paragraph{Reader prompts and portability.}
Each packet is wrapped by three reader prompt families: a plain reader, a suppression reader that ignores updated/corrected/deleted/blocked entries, and a ledger-aware reader that prefers the latest applicable entry. For each question--model pair, the structured evaluation crosses five packet levels, three readers, and two internal backends, yielding $5\times3\times2=30$ structured conditions; Table~\ref{tab:main_results} aggregates over the six reader/backend variants for each packet level. Porting \RENDER to a new memory-QA dataset requires a turn-addressed dialogue adapter, a question-to-slot mapping, and a source of conflict sets. The renderers and reader prompts are otherwise dataset-independent; porting to \LoCoMo{} required about 200 lines of Python and no renderer changes.

\section{Experimental Setup and Results}
\label{sec:exp}

\paragraph{Setup.}
We evaluate \RENDER primarily on the oracle tier of \LongMemEval{}, which contains 500 memory-QA questions with multi-session updates, temporal scope, and abstention~\citep{longmemeval2024}. We test nine commercial models from three providers: \OpenAI{} (\texttt{gpt-5.4-mini}, \texttt{gpt-5.2}, \texttt{gpt-5.2-pro}), \Anthropic{} (\texttt{claude-sonnet-4}, \texttt{claude-opus-4.1}, \texttt{claude-opus-4.6}), and \Google{} (\texttt{gemini-3-flash}, \texttt{gemini-2.5-flash-lite}, \texttt{gemini-3.1-flash-lite}). Exact API IDs are listed in Appendix~\ref{app:model_ids}. Provider-native tools, search, and memory are disabled. The main \LongMemEval{} ladder contains 139{,}500 calls; auxiliary budget, template, matched-content, retrieval-noise, \LoCoMo{}, and \HotpotQA{} experiments plus LLM-judge rescoring bring the study to about 238{,}000 calls.

Our primary metric is accuracy over all questions: a response is correct iff it does not begin with an explicit \texttt{ABSTAIN} refusal and the lowercased gold answer appears as a substring of the lowercased output. Abstentions count as incorrect. Appendix~\ref{app:judge} cross-validates the headline results with normalized exact match and an independent LLM judge. Appendix~\ref{app:abstain-leakage} also audits late-abstention leakage in the substring scorer; a stricter abstention-anywhere check shifts all headline aggregate accuracies by less than 1 point and leaves qualitative conclusions unchanged.

\paragraph{Condition map.}
We use two resolved-$P_2$ conditions for different controls. Table~\ref{tab:main_results} reports the formal packet ladder: $P_2$ uses the formal reader prompt and packet machinery averaged over reader prompts and backends. Table~\ref{tab:token_matched} reports a streamlined resolved-$P_2$ budget control: the resolved value is wrapped in a shorter reader prompt without the formal blocker and abstention fields. Raw baselines also differ by control: Table~\ref{tab:main_results} uses full dialogue in the ladder setup, Table~\ref{tab:token_matched} reports both full and recency-truncated raw under the budget prompt, and Figure~\ref{fig:real_system} uses raw dialogue inside the deployed-template prompt.

\subsection{Accuracy Rises When the Answer Enters the Packet}
\label{sec:stepfunction}

\begin{figure}[t]
\centering
\includegraphics[width=\columnwidth]{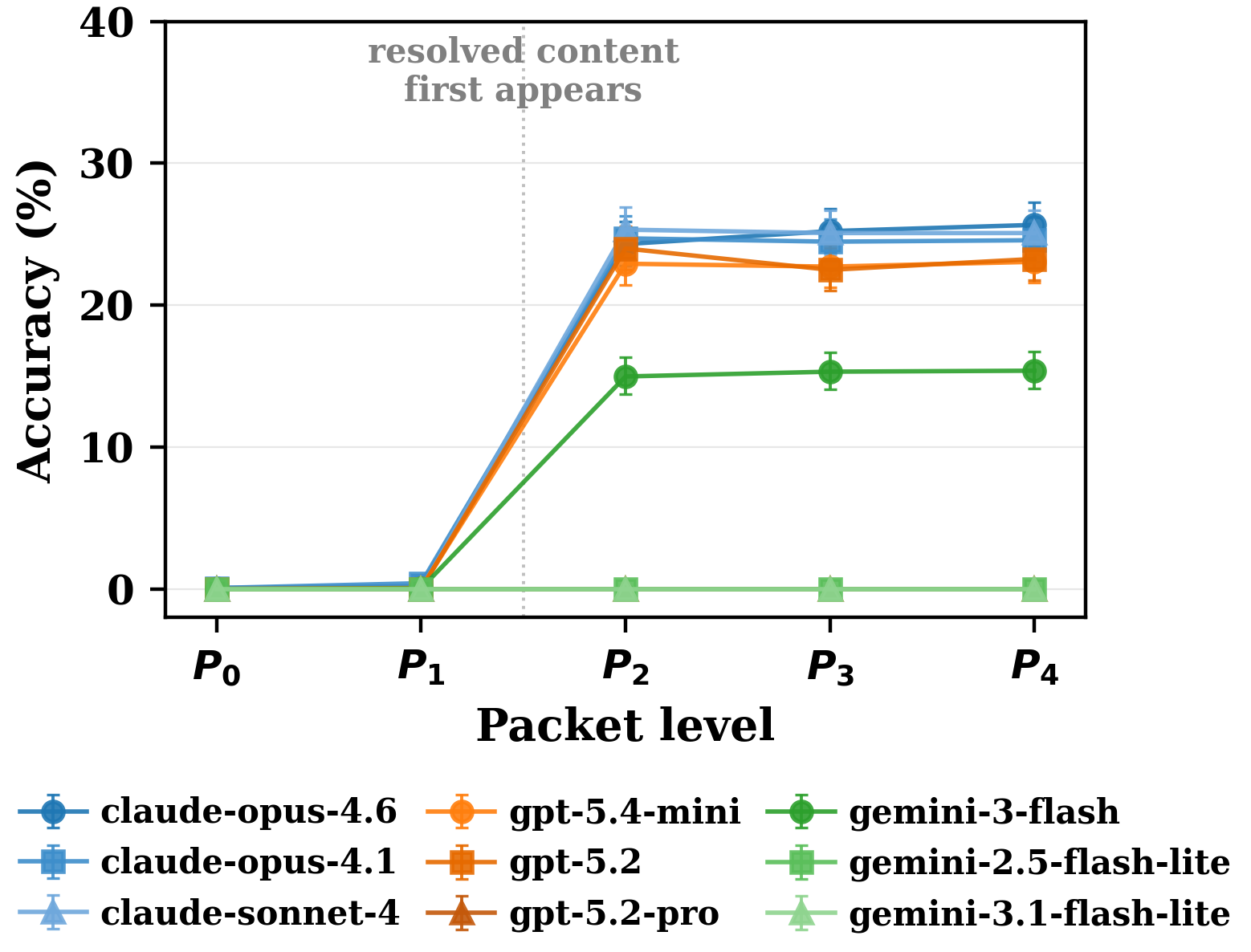}
\caption{Accuracy across the five-packet ladder on \LongMemEval{} (500 questions). $P_0/P_1$ expose witness addresses only; $P_2$ first writes the resolved answer into the packet body; $P_3/P_4$ add metadata around that value.}
\label{fig:stepfunction}
\end{figure}

Figure~\ref{fig:stepfunction} shows the core diagnostic. When packets expose only witness addresses ($P_0/P_1$), all nine models remain near zero accuracy (0.0--0.6\%). Accuracy rises at $P_2$, the first packet level that resolves the conflict set and writes a current-state value into the packet body; the six responsive models reach 15.0--25.3\%. Adding metadata above $P_2$ has little effect: $P_3$/$P_4$ shift accuracy by at most $\pm2$ points per model. Three models remain at 0.0\% across all formal structured packets; Section~\ref{sec:regimes} analyzes these hard-abstention cases.

Full raw conversation is a strong unconstrained baseline because it includes the entire dialogue. It outperforms budgeted formal packets on every model (Table~\ref{tab:main_results}), with raw-over-structured gaps from 10.5 to 44.4 points. This comparison is informative but not causal: it mixes surface with context length, because raw conversation preserves all recency cues and earlier answer-bearing turns while structured packets are budgeted.

\begin{table}[t]
\centering
\small
\setlength{\tabcolsep}{3pt}
\caption{Main \LongMemEval{} results. Raw uses full dialogue; formal packet columns average over 3 reader prompts $\times$ 2 backends per question. $\Delta$ is each row's max--min spread.}
\label{tab:main_results}
\begin{adjustbox}{width=\columnwidth}
\begin{tabular}{@{}lccccccc@{}}
\toprule
Model & Raw & $P_0$ & $P_1$ & $P_2$ & $P_3$ & $P_4$ & $\Delta$ \\
\midrule
\texttt{claude-opus-4.6} & \textbf{45.2} & 0.5 & 0.6 & 24.3 & 24.9 & 25.3 & 44.7 \\
\texttt{claude-opus-4.1} & \textbf{36.8} & 0.4 & 0.6 & 24.7 & 24.0 & 24.1 & 36.4 \\
\texttt{claude-sonnet-4} & \textbf{35.8} & 0.4 & 0.4 & 25.3 & 24.7 & 24.7 & 35.4 \\
\texttt{gpt-5.4-mini} & \textbf{35.4} & 0.4 & 0.5 & 22.9 & 22.4 & 22.7 & 35.0 \\
\texttt{gpt-5.2} & \textbf{42.4} & 0.4 & 0.4 & 24.3 & 22.4 & 23.2 & 42.0 \\
\texttt{gpt-5.2-pro} & \textbf{44.4} & 0.0 & 0.0 & 0.0 & 0.0 & 0.0 & 44.4 \\
\texttt{gemini-3-flash} & \textbf{34.6} & 0.2 & 0.2 & 15.0 & 15.1 & 15.2 & 34.4 \\
\texttt{gemini-2.5-flash-lite} & \textbf{29.2} & 0.0 & 0.0 & 0.0 & 0.0 & 0.0 & 29.2 \\
\texttt{gemini-3.1-flash-lite} & \textbf{38.2} & 0.0 & 0.0 & 0.0 & 0.0 & 0.0 & 38.2 \\
\bottomrule
\end{tabular}
\end{adjustbox}
\end{table}

\subsection{Budget Control Reverses the Raw-vs-Structured Conclusion}
\label{sec:token_matched}

To compare artifacts under similar budgets, we truncate raw conversation to about 800 words, keeping the most recent turns, and compare it with a streamlined resolved-$P_2$ packet on the same 500 questions. Under this control, truncated raw conversation scores only 7.6--12.0\%, while streamlined $P_2$ scores 50.0--82.0\%, yielding a 42.4--72.6 point advantage for structured evidence on every model (Table~\ref{tab:token_matched}). The judge-rescored gap remains positive and significant for every model, ranging from +48.4 to +80.2 points (Appendix~\ref{app:judge-gap}). The result is not that structure is intrinsically better than raw dialogue; rather, once the budget is tight, recency truncation often drops earlier answer-bearing turns, whereas resolved $P_2$ preserves the value regardless of its original position.

\begin{table}[t]
\centering
\small
\setlength{\tabcolsep}{3pt}
\caption{Word-budget comparison on \LongMemEval{} (500 questions). Truncated raw keeps the most recent $\sim$800 words. $\Delta$ = streamlined $P_2$ minus truncated raw; all deltas are significant under paired bootstrap.}
\label{tab:token_matched}
\begin{adjustbox}{width=\columnwidth}
\begin{tabular}{@{}lcccc@{}}
\toprule
Model & Full raw & Trunc. raw & Stream. $P_2$ & $\Delta$ \\
\midrule
\texttt{claude-opus-4.6} & 51.4 & 10.4 & \textbf{76.0} & $+65.6$ \\
\texttt{claude-opus-4.1} & 40.4 & 11.0 & \textbf{72.2} & $+61.2$ \\
\texttt{claude-sonnet-4} & 33.2 & 10.2 & \textbf{78.4} & $+68.2$ \\
\texttt{gpt-5.4-mini} & 38.8 & 11.6 & \textbf{73.0} & $+61.4$ \\
\texttt{gpt-5.2} & 39.4 & 11.0 & \textbf{77.0} & $+66.0$ \\
\texttt{gpt-5.2-pro} & 50.4 & 12.0 & \textbf{77.0} & $+65.0$ \\
\texttt{gemini-3-flash} & 30.4 & 7.6 & \textbf{50.0} & $+42.4$ \\
\texttt{gemini-2.5-flash-lite} & 25.4 & 9.4 & \textbf{57.2} & $+47.8$ \\
\texttt{gemini-3.1-flash-lite} & 37.2 & 9.4 & \textbf{82.0} & $+72.6$ \\
\bottomrule
\end{tabular}
\end{adjustbox}
\end{table}

This comparison also clarifies hard abstention. In the main ladder, \texttt{gpt-5.2-pro} scores 0.0\% on every formal \RENDER packet; in the budget-matched condition, it scores 77.0\% on the same active-only packet body when wrapped in a streamlined instruction prompt rather than a formal reader prompt with explicit blocker and abstention machinery. The same pattern holds for the other hard-abstention models, so we treat the refusal result as an artifact-and-prompt interaction rather than a pure field-format effect.

\subsection{Deployed Memory Templates Change Scores by up to 49 Points}
\label{sec:real_system}

We next approximate reader-facing outputs from deployed memory strategies using deterministic template families: \ChatGPT{}-style natural-language entries, \LangChain{}-style session summaries, \MemGPT{}-style typed JSON records, and raw conversation. This is a deployed-template comparison rather than a pure same-content ablation: \ChatGPT{} and \MemGPT{}-style templates receive up to five gold-localized turns, \LangChain{} summaries truncate each session to its first few turns, and raw conversation includes the full dialogue.

\begin{figure}[t]
\centering
\includegraphics[width=\columnwidth]{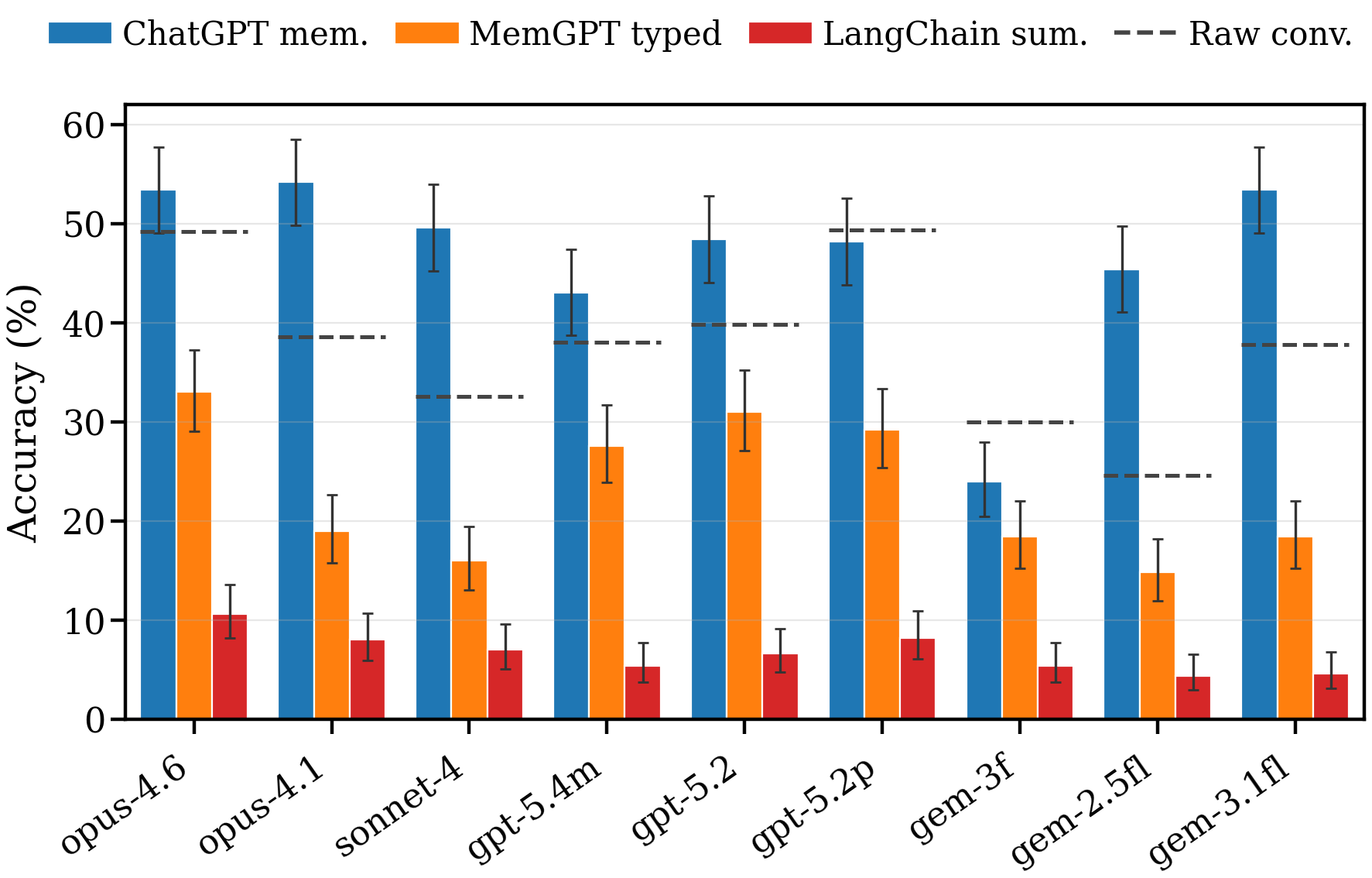}
\caption{Deployed-template comparison on \LongMemEval{} (500 questions). Under the primary scorer, \ChatGPT{}-style entries have higher point estimates than raw conversation on 7 of 9 models; judge rescoring shows mixed model-specific significance.}
\label{fig:real_system}
\end{figure}

Surface choice substantially changes the reported score (Figure~\ref{fig:real_system}). Under the primary substring scorer, \ChatGPT{}-style entries have higher point estimates than raw conversation on seven of nine models, with gains of 4.2--20.8 points; the two exceptions are \texttt{gemini-3-flash} ($-6.0$) and \texttt{gpt-5.2-pro} ($-1.2$). Judge rescoring preserves a positive aggregate effect (+7.9pp) but shows mixed model-specific significance (Appendix~\ref{app:judge-chatgpt}). \MemGPT{}-style typed records are model-dependent (14.8--33.0\%), while \LangChain{} summaries are consistently weakest (4.4--10.6\%). Across models, the best--worst surface spread ranges from 24.6 to 48.8 points. The three models that score 0.0\% on formal \RENDER packets answer the same facts from \ChatGPT{}-style entries at 45.4--53.4\%, indicating that formal ledger packets can trigger refusal even when compact natural-language evidence remains usable.

\subsection{Generalization and Robustness}
\label{sec:robustness}
\label{sec:retrieval_noise}
\label{sec:format_transfer}

The pattern is not specific to the main \LongMemEval{} setting. On 200 \LoCoMo{} questions, every model scores at most 0.5\% when only witness addresses are visible and recovers to 32.0--97.0\% at active $P_2$; no model hard-abstains across all structured conditions (Appendix~\ref{app:additional_diagnostics}). Under retrieval noise, aggregate raw accuracy drops 8.8 points from $k{=}0$ to $k{=}9$ distractors, while \ChatGPT{}-style entries drop only 1.5 points; the NL-over-raw advantage shifts from $-1.5$ to $+5.9$ points, with per-model heterogeneity (Appendix~\ref{app:retrieval_noise}). On \HotpotQA{}, a non-conversational multi-hop QA task, aggregate accuracy follows NL entries (69.8\%) $>$ raw paragraphs (68.6\%) $>$ summaries (67.2\%) $>$ typed records (62.7\%), suggesting transfer beyond memory QA while also confirming that surface preferences are task-dependent (Appendix~\ref{app:hotpot_transfer}).

\section{Analysis}
\label{sec:analysis}

\subsection{Representation-induced behavior regimes}
\label{sec:regimes}

Reader-facing evidence artifacts change answer behavior as well as accuracy. The answer-rate drop from raw conversation to \texttt{full\_render} ranges from 28.1 to 75.2 points, splitting models into three regimes (visualized in Appendix~\ref{app:additional_diagnostics}). \emph{Hard abstention}: \texttt{gpt-5.2-pro}, \texttt{gemini-2.5-flash-lite}, and \texttt{gemini-3.1-flash-lite} have 0.0\% answer rate on every formal packet, despite answering from raw conversation, \ChatGPT{}-style memory entries (45.4--53.4\%), and unresolved witness text (18--19\%). \emph{Cautious abstention}: \texttt{claude-opus-4.6}, \texttt{claude-opus-4.1}, \texttt{claude-sonnet-4}, \texttt{gpt-5.4-mini}, and \texttt{gpt-5.2} abstain on $P_0/P_1$ but answer normally once $P_2$ resolves the conflict. \emph{Speculative answering}: \texttt{gemini-3-flash} answers 22\% of unresolved packets at 0.0\% accuracy. These regimes do not track provider or model generation.

\subsection{Why we report overall accuracy}

Answered-only accuracy can reward abstention. \texttt{claude-sonnet-4} scores 24.7\% overall under \texttt{full\_render} at a 29.5\% answer rate, yielding 83.7\% conditional accuracy mostly because it abstains. \texttt{gemini-3-flash} shows the opposite pattern, answering 43.9\% of items at 15.2\% overall. We therefore report correct answers divided by total questions, with abstentions counted as incorrect.

\section{Discussion}
\label{sec:discussion}
\label{sec:mechanism-hypotheses}

\paragraph{What drives the artifact effect.}
The $P_1{\to}P_2$ jump bundles two operations: the reader gains answer-bearing content, and the upstream system has resolved competing states. A matched-content control separates them by quoting the witness turn verbatim while withholding explicit slot resolution and the \texttt{active\_value} field (Appendix~\ref{app:additional_diagnostics}). Content visibility is the largest driver in this control: witness text alone lifts accuracy from near zero to 14.2--23.2\%. For the six responsive models, explicit conflict resolution adds a smaller 0.8--6.1 points, significant for three. The hard-abstention models sharpen the conclusion: they answer 18--19\% of unresolved witness-text items correctly but 0\% of resolved formal packets, consistent with artifact- or prompt-induced refusal rather than inability to reason from structured evidence.

Full raw conversation still helps when context is unconstrained because it preserves recency ordering, update phrasing, and local coherence. Under a tight budget, however, recency truncation can remove earlier answer-bearing turns, while a resolved $P_2$ packet preserves the answer. Compact natural-language memory entries benefit from both properties: they retain enough discourse texture for fluent reading while remaining short enough to include the relevant fact.

\paragraph{Using \RENDER as a benchmark control.}
\RENDER is not intended to identify a universal best representation. Its purpose is to make the reader-facing evidence surface an explicit experimental variable. Benchmark authors can evaluate each system's native artifact alongside a fixed-surface control, such as a resolved $P_2$ packet or deterministic NL entry. System builders can use the packet ladder to localize failure: near-zero $P_0/P_1$ followed by recovery at $P_2$ indicates that answer-bearing content is not reaching the reader; persistently low $P_4$ accuracy points to surface-induced refusal or format misinterpretation. Deployment teams can keep structured storage internally for auditability, deduplication, and retrieval while rendering compact natural language for the reader.

\paragraph{Candidate mechanisms and limits of the hierarchy.}
Three mechanisms are consistent with the data. Compact NL entries may better match the text distribution frontier models are trained on than typed records or formal ledgers. Formal packets include fields such as \texttt{blocker}, \texttt{active\_value}, and \texttt{should\_abstain}, which may interact with instruction-following or safety-tuned refusal heuristics. Raw dialogue carries discourse cues that typed rendering strips away, while truncation removes earlier evidence entirely. These mechanisms are not mutually exclusive, and the preferred surface is task- and model-dependent: NL entries lead deployed templates on \LongMemEval{}, but raw Wikipedia paragraphs beat NL entries for two \Anthropic{} readers on \HotpotQA{}, and \LangChain{} summaries outperform typed triples on \HotpotQA{} despite being weakest on \LongMemEval{}. The practical implication is that the raw-vs-structured dichotomy is too coarse; compactness, naturalness, metadata density, and abstention signaling all matter.

\section{Conclusion}
\label{sec:conclusion}

The text handed to the reader by a memory/RAG layer is not always a neutral implementation detail. Across nine models, two memory benchmarks, \HotpotQA{}, and about 238{,}000 model calls, the same underlying questions produce very different accuracies depending on the reader-facing evidence artifact. Full raw conversation is a strong baseline when context is unconstrained, but under matched word budgets a streamlined resolved packet beats recency-truncated raw dialogue by 42.4--72.6 points. In deployed-style templates, \ChatGPT{}-style memory entries have higher primary-scorer point estimates than raw conversation on 7 of 9 models, while judge rescoring preserves a positive aggregate effect but shows mixed model-specific significance. \LangChain{} summaries are weakest, and formal ledger packets plus abstention machinery trigger near-complete abstention in 3 of 9 models even though those same models answer natural-language memory entries at 45.4--53.4\%.

The result does not imply that structure is always bad or that raw dialogue is always best. Evidence rendering is an uncontrolled experimental variable unless it is reported or fixed. \RENDER turns that variable into a control: benchmark authors can add a fixed evidence-surface condition, system builders can distinguish missing content from unresolved conflict or refusal-triggering surfaces, and deployment teams can keep structured storage while rendering compact natural language to the reader.

\section{Limitations}
\label{sec:limitations}

\paragraph{Design and measurement choices.}
The token-matched baseline uses word-level truncation as a proxy for token-budget matching; provider-specific tokenizers would be a tighter control, and recency-preserving truncation can miss earlier answer-bearing content. The budgeted comparison therefore does not isolate oracle evidence localization from surface form; an oracle raw-excerpt baseline would further separate localization from rendering. The deployed-template comparison (\S\ref{sec:real_system}) uses deterministic templates that mimic deployed outputs rather than running the systems end to end, so it omits real retrieval errors, summarization errors, and pipeline artifacts. The substring scorer can over-credit verbose answers and miss semantic paraphrases; on \LoCoMo{} it also penalizes raw-conversation answers that express dates relatively. Appendix~\ref{app:judge} cross-validates the headline ordering with normalized exact match and an independent LLM judge, but smaller metadata and formality effects remain more sensitive to scoring choices and sampling noise than the headline step function, token-matched gap, and deployed-template spread.

\paragraph{Template and retrieval scope.}
\RENDER deliberately controls the artifact shown to the reader after evidence has been selected. This design is useful for isolating rendering effects, but it does not measure the full quality of a memory or RAG pipeline. A production system must still decide what to store, when to update stale facts, how to retrieve relevant memories, and how to suppress irrelevant or sensitive material. Our deployed-template variants therefore should be read as controlled approximations of common reader inputs, not as evaluations of \ChatGPT{}, \LangChain{}, \MemGPT{}, or any other product implementation. In particular, \ChatGPT{}-style and \MemGPT{}-style renderings receive gold-localized turns, while \LangChain{}-style summaries use fixed deterministic truncation. This makes the comparison transparent and repeatable, but it also means that absolute accuracies should not be interpreted as production-system leaderboard scores.

\paragraph{Boundary with structured generation.}
\RENDER does not study whether requiring a model to output JSON, XML, schemas, or other structured formats harms reasoning. The answer contract is fixed across conditions; only the evidence representation changes before the reader answers. The results are therefore not an output-side format-tax measurement and are not evidence about constrained decoding.

\paragraph{Scope and external validity.}
Evidence comes from \LongMemEval{}'s oracle tier, where conflict sets are annotated rather than produced by a retriever. The retrieval-noise experiment uses random distractors and is only a partial substitute for learned retriever error. We evaluate conversational memory and multi-hop Wikipedia QA, not document-grounded RAG or multimodal settings. The nine models are commercial APIs whose system prompts, safety policies, and calibration we cannot inspect; evaluations ran in April 2026 using the exact API IDs listed in Appendix~\ref{app:model_ids}. The \HotpotQA{} summary and typed surfaces are generated by \texttt{claude-opus-4.6}, which is also one evaluator; Appendix~\ref{app:hotpot_transfer} checks this possible generator/reader overlap and finds no systematic advantage for that model. The \ChatGPT{}-style memory template exposes both old and new values, which helps conflict resolution; a production memory store keeping only the latest value would face different tradeoffs. Three models hard-abstain on formal \RENDER packets while answering normally on NL entries and unresolved witness text, but we do not identify whether refusal is driven by field names, explicit abstention cues, ledger terminology, or the formal reader prompt.

\paragraph{Statistical interpretation.}
The largest effects in the paper are robust across scoring variants: answer-bearing $P_2$ packets beat recency-truncated raw dialogue for every model, and the aggregate deployed-template spread is large under both substring and judge scoring. Some finer-grained claims are intentionally more cautious. For example, \ChatGPT{}-style entries have higher primary-scorer point estimates than raw conversation on 7 of 9 models, but judge rescoring shows mixed model-specific significance. We therefore treat those results as evidence that rendering can materially change measured accuracy, not as a universal ranking of one surface over another.

\paragraph{Implementation heterogeneity for \texttt{gpt-5.2-pro}.}
\OpenAI{} exposes \texttt{gpt-5.2-pro} only through the \texttt{v1/responses} endpoint, not the chat-completions surface used for the other two \OpenAI{} models. Both endpoints receive identical prompt strings at temperature 0 with tools and provider-native memory disabled, but the endpoint difference remains a confound for any \texttt{gpt-5.2-pro} comparison.

\section{Ethical Considerations}
\label{sec:ethics}

\RENDER studies evaluation artifacts, not deployed user-facing memory behavior. The experiments use released benchmark data and do not introduce new human-subject data collection. Because conversational memory and RAG systems can surface sensitive or stale personal information, evidence rendering should be reported and controlled, not treated as a recommendation to expose raw private histories. The released artifacts support audit, not deployment safety claims; production systems still require privacy review, data minimization, consent, access controls, and task-specific refusal testing.

\paragraph{Release boundaries.}
The released artifacts support scientific audit of the measurements. They contain rendered benchmark artifacts, auxiliary-experiment model responses, scorer outputs, stored judge verdicts, main-ladder aggregate summaries, and offline verification scripts, but they are not a deployment recipe. We do not release private user data, credentials, provider logs with account identifiers, or provider-calling runner scripts. Downstream users should respect the source dataset licenses and avoid reidentification attempts.

\paragraph{Interpreting refusal behavior.}
One ethical risk is over-reading the hard-abstention result. A model that refuses a formal packet here is not necessarily safer in deployment, and a model that answers compact natural-language evidence is not necessarily unsafe. Refusal depends on instructions, provider policy, and application context; deployment teams should validate behavior on their own task distributions rather than transferring the benchmark result directly.

\bibliography{references}

\clearpage
\appendix

\section{Model and Run Details}
\label{app:model_ids}

All calls used temperature~0 and had provider-native tools, search, and memory disabled. Max output tokens were 128 for reader calls except \texttt{gpt-5.2-pro}, which used the \texttt{v1/responses} endpoint with 1024 tokens. Table~\ref{tab:model_ids} maps the table-facing model names to the exact API IDs used in the runs; provider documentation records model-code and versioning conventions~\citep{openai2026models,anthropic2026modelids,google2026geminimodels}.

\begin{table}[h]
\centering
\scriptsize
\setlength{\tabcolsep}{2pt}
\caption{Display names used in result tables and exact API IDs used in the runs.}
\label{tab:model_ids}
\begin{tabularx}{\columnwidth}{@{}lX@{}}
\toprule
Display name & Exact API ID \\
\midrule
\texttt{gpt-5.4-mini} & \path{gpt-5.4-mini-2026-03-17} \\
\texttt{gpt-5.2} & \path{gpt-5.2-2025-12-11} \\
\texttt{gpt-5.2-pro} & \path{gpt-5.2-pro-2025-12-11} \\
\texttt{claude-sonnet-4} & \path{claude-sonnet-4-20250514} \\
\texttt{claude-opus-4.1} & \path{claude-opus-4-1-20250805} \\
\texttt{claude-opus-4.6} & \path{claude-opus-4-6} \\
\texttt{gemini-3-flash} & \path{gemini-3-flash-preview} \\
\texttt{gemini-2.5-flash-lite} & \path{gemini-2.5-flash-lite} \\
\texttt{gemini-3.1-flash-lite} & \path{gemini-3.1-flash-lite-preview} \\
\bottomrule
\end{tabularx}
\end{table}

\section{Additional Diagnostics and Replications}
\label{app:additional_diagnostics}

\subsection{Representation-induced behavior regimes}
\label{app:regimes}

\begin{figure}[t]
\centering
\includegraphics[width=\columnwidth]{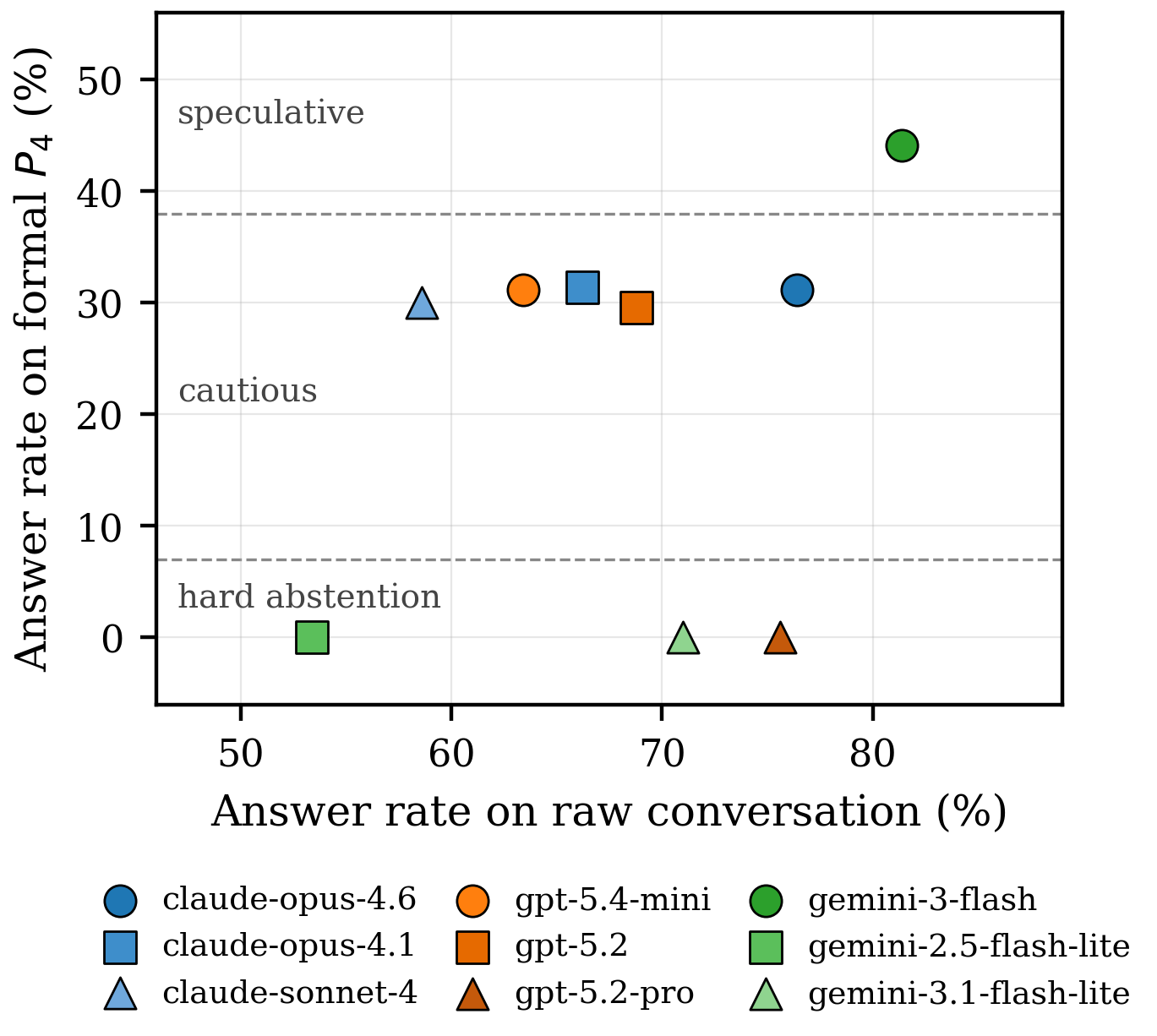}
\caption{Representation-induced behavior regimes. Each dot is one model, plotted by answer rate on raw conversation (x) vs. answer rate on the formal $P_4$ packet (y).}
\label{fig:regimes}
\end{figure}

Figure~\ref{fig:regimes} visualizes the answer-rate regimes discussed in \S\ref{sec:regimes}. The hard-abstention models have nonzero answer rates on raw conversation but near-zero answer rates on formal $P_4$ packets.

\subsection{\LoCoMo{} packet-ladder replication}
\label{app:locomo}

The packet-ladder effect replicates on \LoCoMo{}: $P_0$ values are 0.0--0.5\%, while answer-bearing $P_2$ packets recover strongly (Table~\ref{tab:locomo}). Active-only accuracy exceeds raw-conversation accuracy here, but the substring scorer penalizes raw answers that express dates relatively; we interpret this table as a diagnostic replication rather than a general raw-vs-structured ranking.

\begin{table}[t]
\centering
\small
\setlength{\tabcolsep}{5pt}
\caption{Replication on \LoCoMo{} (200 questions). $P_0$ values, omitted for compactness, are 0.0--0.5\%.}
\label{tab:locomo}
\begin{tabular}{@{}lcc@{}}
\toprule
Model & Raw conv. & $P_2$ active \\
\midrule
\texttt{claude-opus-4.6} & 22.5 & 96.0 \\
\texttt{claude-opus-4.1} & 21.0 & 96.0 \\
\texttt{claude-sonnet-4} & 19.5 & 97.0 \\
\texttt{gpt-5.4-mini} & 18.0 & 94.5 \\
\texttt{gpt-5.2} & 21.5 & 83.5 \\
\texttt{gpt-5.2-pro} & 20.0 & 86.5 \\
\texttt{gemini-3-flash} & 16.0 & 32.0 \\
\texttt{gemini-2.5-flash-lite} & 18.5 & 82.5 \\
\texttt{gemini-3.1-flash-lite} & 17.5 & 94.5 \\
\bottomrule
\end{tabular}
\end{table}

\subsection{Matched-content control}
\label{app:matched_content}

Table~\ref{tab:matched_content} separates content exposure from explicit conflict resolution. Witness text quotes the relevant turn verbatim but withholds the resolved slot and \texttt{active\_value} field.

\begin{table}[t]
\centering
\small
\setlength{\tabcolsep}{5pt}
\caption{Matched-content control on \LongMemEval{} (500 questions). Bold marks models that answer from witness text yet hard-abstain on resolved formal packets.}
\label{tab:matched_content}
\begin{tabular}{@{}lcc@{}}
\toprule
Model & Witness text & Active $P_2$ \\
\midrule
\texttt{claude-opus-4.6}       & 23.2          & 24.3 \\
\texttt{claude-opus-4.1}       & 18.6          & 24.7 \\
\texttt{claude-sonnet-4}       & 19.6          & 25.3 \\
\texttt{gpt-5.4-mini}          & 21.6          & 22.9 \\
\texttt{gpt-5.2}               & 18.6          & 24.3 \\
\texttt{gpt-5.2-pro}           & \textbf{18.0} & 0.0  \\
\texttt{gemini-3-flash}        & 14.2          & 15.0 \\
\texttt{gemini-2.5-flash-lite} & \textbf{19.4} & 0.0  \\
\texttt{gemini-3.1-flash-lite} & \textbf{19.2} & 0.0  \\
\bottomrule
\end{tabular}
\end{table}

\section{Robustness: LLM-Judge Rescoring}
\label{app:judge}

Substring matching is a coarse scorer. We cross-validate it against two stronger signals: a rule-based normalized-EM baseline and an LLM judge.

\subsection{Methodology}
\label{app:judge-method}

\paragraph{Full-population judge.}
The 31{,}500 (model, question, condition) triples in the token-matched (13{,}500) and real-system (18{,}000) experiments are rescored by \texttt{claude-haiku-4-5-20251001} at temperature 0 with a 4-token cap. The judge sees only the gold answer and the model response and outputs \texttt{CORRECT} or \texttt{INCORRECT}. All judge calls completed successfully. The exact system prompt is:

\begin{quote}\small
\noindent You grade whether a model response correctly answers a factual question, given the gold answer. Output exactly one word: \texttt{CORRECT} or \texttt{INCORRECT}.

\smallskip
\noindent Rules:
\begin{itemize}[leftmargin=1.2em,itemsep=1pt,topsep=1pt]
\item Accept semantic paraphrases that preserve the required information (e.g., ``Denver'' matches ``Denver, CO''; ``Jan 2024'' matches ``January 2024'').
\item Accept normalized dates, numbers, and names when the referent is unambiguous.
\item Reject abstentions or refusals (e.g., ``ABSTAIN'', ``I don't know'').
\item Reject responses that omit required specificity (e.g., gold ``John Smith'' vs.\ response ``John'' is \texttt{INCORRECT}).
\item Reject hallucinations or answers that contradict the gold.
\item Be strict: when in doubt, prefer \texttt{INCORRECT}.
\end{itemize}
\end{quote}

The archive includes offline judge-audit and aggregation scripts: \texttt{scripts/rescore\_with\_judge.py} and \texttt{scripts/aggregate\_full\_judge.py}. These scripts recompute scorer agreement, judge aggregates, and bootstrap intervals from the stored verdicts without making provider calls.

\paragraph{Second-judge cross-validation.}
A stratified 600-item sample across all three headline experiments (including the 4{,}500-item matched-content control) is rescored by an independent second judge with the same prompt. The primary-vs-second comparison uses the 516 token-matched and real-system items with non-empty response text in both judge logs; matched-content items appear only in the stratified scorer columns because they are not part of the full-population judge pool. On this cleaned overlap the two judges agree on \textbf{90.7\%} of items (92.2\% real-system, 88.8\% token-matched).

\paragraph{Normalized exact match.}
Rule-based baseline: lowercase, strip punctuation, drop articles, token-level substring containment in either direction, reject abstention tokens.

\subsection{Agreement rates}
\label{app:judge-agreement}

Table~\ref{tab:agreement} gives pairwise agreement between scorers. On the full population, substring and the primary judge agree on 87--88\% of items and normalized-EM and the primary judge agree on 89--90\%. On the cleaned 516-item overlap, the two judges agree on 90.7\%.

\begin{table}[!t]
\centering
\small
\setlength{\tabcolsep}{4pt}
\caption{Scorer agreement. The TM and RS columns report agreement with the \emph{primary} judge on the full-population verdicts; the \emph{Strat.}\ column reports agreement with the \emph{second} judge on the 600-item stratified sample. The final row reports primary-vs-second-judge agreement on the cleaned 516-item overlap shared with the token-matched and real-system full-population rescores.}
\label{tab:agreement}
\begin{tabular}{@{}lccc@{}}
\toprule
Comparison & TM & RS & Strat. \\
 & \footnotesize(13.5K) & \footnotesize(18K) & \footnotesize(600) \\
\midrule
sub.\ vs.\ prim.\ judge   & 87.2 & 87.9 & 68.8 \\
norm.\ vs.\ prim.\ judge  & 89.5 & 89.5 & 72.8 \\
sub.\ vs.\ norm.\         & 96.5 & 95.8 & 56.3 \\
prim.\ vs.\ 2nd judge     & 88.8 & 92.2 & 90.7$^\dagger$ \\
\bottomrule
\end{tabular}

\vspace{2pt}
\footnotesize $^\dagger$on cleaned 516-item overlap; TM = token-matched, RS = real-system; all values \%.
\end{table}

\subsection{Per-condition rescored accuracy}
\label{app:judge-percond}

Table~\ref{tab:rescore_condition} gives per-condition accuracy under all three scorers, aggregated across nine models. Judge accuracy is 3.8--13.3pp higher than substring on every condition; the relative ordering within each experiment is preserved. The judge is more permissive than either rule-based scorer because it accepts semantic paraphrases.

\begin{table}[!t]
\centering
\small
\setlength{\tabcolsep}{3pt}
\caption{Per-condition accuracy (\%) under three scorers, aggregated across nine models. Full-population rescore for TM and RS; 600-item stratified rescore reweighted to population for matched-content. The matched-content row reports the analyzed row count after rescore filtering, hence \(n{=}4{,}423\) rather than 4{,}500.}
\label{tab:rescore_condition}
\begin{tabular}{@{}lrrrr@{}}
\toprule
Condition & $n$ & Sub. & Norm. & Judge \\
\midrule
\multicolumn{5}{@{}l}{\textit{token-matched}} \\
Full raw         & 4{,}500 & 38.5 & 38.6 & 50.5 \\
Trunc.\ raw      & 4{,}500 & 10.3 & 11.3 & 15.2 \\
Active $P_2$     & 4{,}500 & 71.4 & 73.0 & \textbf{84.5} \\
\midrule
\multicolumn{5}{@{}l}{\textit{real-system}} \\
\ChatGPT{}-mem      & 4{,}500 & 46.6 & 46.2 & \textbf{57.9} \\
\LangChain sum.   & 4{,}500 &  6.7 &  6.9 &  9.7 \\
\MemGPT typed     & 4{,}500 & 23.0 & 25.9 & 34.1 \\
Raw conv.        & 4{,}500 & 37.8 & 38.2 & 50.0 \\
\midrule
\multicolumn{5}{@{}l}{\textit{matched-content}} \\
Witness text     & 4{,}423 & 19.5 & 20.1 & 24.0 \\
\bottomrule
\end{tabular}
\end{table}

\subsection{Headline token-matched gap: per-model CIs}
\label{app:judge-gap}

Table~\ref{tab:rescore_gap} reports the per-model $P_2 - $ truncated-raw gap under substring and under the Claude Haiku judge, with paired bootstrap 95\% CIs (10{,}000 resamples over the 500 shared questions). Every model's gap is positive and significant under both scorers. The judge gap ranges $+$48.4 to $+$80.2pp, and each per-model judge gap exceeds the corresponding substring gap; the 42.4--72.6pp substring range is a conservative lower bound relative to the semantic judge.

\begin{table}[!t]
\centering
\small
\setlength{\tabcolsep}{3pt}
\caption{Token-matched gap ($P_2 - $ trunc.\ raw) per model, under substring (original) and the Claude Haiku judge with paired bootstrap 95\% CIs over 500 questions. No CI contains zero.}
\label{tab:rescore_gap}
\begin{tabular}{@{}lrrr@{}}
\toprule
Model & Substr. & Judge & 95\% CI \\
\midrule
\texttt{opus-4.6}    & +65.6 & +80.2 & [76.6, 83.6] \\
\texttt{opus-4.1}    & +61.2 & +70.2 & [66.0, 74.4] \\
\texttt{sonnet-4}    & +68.2 & +76.6 & [72.6, 80.4] \\
\texttt{gpt-5.4-mini}   & +61.4 & +74.0 & [70.0, 77.8] \\
\texttt{gpt-5.2}        & +66.0 & +75.2 & [71.2, 79.0] \\
\texttt{gpt-5.2-pro}    & +65.0 & +73.8 & [70.0, 77.6] \\
\texttt{gem-3-flash}    & +42.4 & +48.4 & [43.6, 53.0] \\
\texttt{gem-2.5-lite}   & +47.8 & +48.8 & [44.2, 53.4] \\
\texttt{gem-3.1-lite}   & +72.6 & +76.4 & [72.6, 80.0] \\
\midrule
Range & 42.4--72.6 & 48.4--80.2 & -- \\
\bottomrule
\end{tabular}
\end{table}

\subsection{Real-system \ChatGPT{}-style vs.\ raw advantage: per-model CIs}
\label{app:judge-chatgpt}

Table~\ref{tab:rescore_chatgpt_raw} reports the real-system \ChatGPT{}-style memory vs.\ raw-conversation comparison by model. Under the judge: 3 significant positive (\texttt{opus-4.1}, \texttt{sonnet-4}, \texttt{gemini-2.5-flash-lite}), 2 significant negative (\texttt{gpt-5.2-pro}, \texttt{gemini-3-flash}), 4 within sampling noise.

\begin{table}[!t]
\centering
\small
\setlength{\tabcolsep}{4pt}
\caption{Real-system \ChatGPT{}-style memory $-$ raw conversation per model, under substring and under the Claude Haiku judge, with paired bootstrap 95\% CIs. Under the judge, the advantage is significantly positive on 3/9 models and significantly negative on 2/9.}
\label{tab:rescore_chatgpt_raw}
\begin{tabular}{@{}lrrr@{}}
\toprule
Model & Substr. & Judge & 95\% CI \\
\midrule
\texttt{opus-4.6}   & +4.2 & +5.6 & [$-$0.2, +11.6] \\
\texttt{opus-4.1}   & +15.6 & +21.2 & [+15.2, +27.2] \\
\texttt{sonnet-4}   & +17.0 & +24.0 & [+17.6, +30.2] \\
\texttt{gpt-5.4-mini} & +5.0 & +4.2 & [$-$2.0, +10.4] \\
\texttt{gpt-5.2}    & +8.6 & +5.6 & [$-$0.6, +11.8] \\
\texttt{gpt-5.2-pro} & $-$1.2 & $-$8.0 & [$-$13.8, $-$2.2] \\
\texttt{gem-3-flash} & $-$6.0 & $-$8.2 & [$-$13.6, $-$2.6] \\
\texttt{gem-2.5-lite} & +20.8 & +20.6 & [+14.2, +27.2] \\
\texttt{gem-3.1-lite} & +15.6 & +6.0 & [$-$0.2, +12.4] \\
\midrule
Aggregate & +7.9 & +7.9 & -- \\
\bottomrule
\end{tabular}
\end{table}

\subsection{Late-abstention and substring-leakage cases in the primary substring scorer}
\label{app:abstain-leakage}

The run-time scorer detects \texttt{ABSTAIN} via a \emph{prefix} check on the response (equivalent to ``starts with \texttt{ABSTAIN}'' for most scripts, or ``\texttt{ABSTAIN} appears in the first 20 characters'' for the \HotpotQA{} scorer). A response that first describes why it cannot answer, mentions the gold string in that explanation, and appends \texttt{ABSTAIN} at the end passes both clauses of the scorer and is counted correct. Three illustrative cases from the released archive:

\begin{itemize}
\item \texttt{retrieval\_noise/results.json}, \texttt{anthropic/claude-opus-4.6}, key \texttt{gpt4\_2312f94c\textbar{}chatgpt\_mem\textbar{}k0}. Gold = ``Samsung Galaxy S22.'' Response: ``Based on the saved memories, I cannot determine which device you got first\ldots{} ABSTAIN.'' Stored as \texttt{correct=true}.
\item \texttt{retrieval\_noise/results.json}, same model, key \texttt{gpt4\_2312f94c\textbar{}raw\textbar{}k3}. Gold = ``Samsung Galaxy S22.'' Response: ``Based on the conversations, the Dell XPS 13 was obtained first\ldots{}'' (no \texttt{ABSTAIN}, but the substantive answer is the \emph{other} device; the gold string appears later in the response's supporting context). Stored as \texttt{correct=true}.
\item \texttt{real\_system/results.json}, \texttt{anthropic/claude-sonnet-4}, qid \texttt{a2f3aa27}, condition \texttt{c2\_langchain\_summary}. Gold = ``1300.'' Response: ``You were nearing 1300 followers on Instagram as of May 28, 2023. However, I cannot determine your exact current follower count from these summaries.'' Stored as \texttt{correct=true}.
\end{itemize}

Tightening the abstention check to \texttt{``ABSTAIN'' in response.upper()} (detect anywhere rather than only at the prefix) reclassifies 1 to 30 items per condition across the headline experiments and shifts aggregate accuracies as follows. This stricter \texttt{ABSTAIN} check addresses late-abstention cases, but it does not remove all substring false positives without an explicit abstention token:

\begin{table}[!t]
\centering
\small
\setlength{\tabcolsep}{3pt}
\caption{Aggregate accuracy under the primary scorer (prefix-only abstain) vs.\ a strict scorer (\texttt{ABSTAIN} anywhere). $\Delta$ = strict $-$ primary, in pp.}
\label{tab:strict_scorer}
\begin{tabular}{@{}lrrr@{}}
\toprule
Condition & Prim. & Strict & $\Delta$ \\
\midrule
\multicolumn{4}{@{}l}{\textit{token-matched}} \\
\texttt{c1\_full\_raw}     & 38.51 & 38.09 & $-0.42$ \\
\texttt{c2\_truncated}     & 10.29 & 10.22 & $-0.07$ \\
\texttt{c3\_active}        & 71.42 & 71.36 & $-0.07$ \\
\midrule
\multicolumn{4}{@{}l}{\textit{real-system}} \\
\ChatGPT{}-style memory & 46.62 & 46.56 & $-0.07$ \\
\LangChain{}-summary  &  6.69 &  6.56 & $-0.13$ \\
\MemGPT{}-typed       & 23.04 & 22.91 & $-0.13$ \\
Raw conversation   & 37.78 & 37.18 & $-0.60$ \\
\midrule
\multicolumn{4}{@{}l}{\textit{retrieval-noise $k{=}9$ agg.}} \\
raw                & 36.06 & 35.61 & $-0.44$ \\
NL (\ChatGPT{}-mem)   & 41.94 & 41.83 & $-0.11$ \\
\bottomrule
\end{tabular}
\end{table}

For \HotpotQA{}, each aggregate surface changes by at most 0.72pp under the strict scorer. The token-matched headline gap shifts negligibly ($+61.1$ vs.\ $+61.1$pp in aggregate; no per-model sign change). The real-system \ChatGPT-minus-raw aggregate is essentially unchanged (+8.84 vs.\ +9.38pp). The \texttt{c1\_full\_raw} and \texttt{c4\_raw\_conversation} conditions take the largest hits because long raw-conversation responses have more chances to mention the gold string before abstaining, but the effect is under 1pp. No qualitative headline conclusion changes under the strict scorer.

\section{Retrieval Noise: Per-Model Detail}
\label{app:retrieval_noise}

Section~\ref{sec:retrieval_noise} reports the aggregate retrieval-noise result; per-model breakdowns and paired bootstrap 95\% CIs are below.

\begin{table*}[!htbp]
\centering
\small
\setlength{\tabcolsep}{4pt}
\caption{Per-model accuracy under retrieval noise, 200 \LongMemEval questions $\times$ 9 models $\times$ 2 surfaces $\times$ 3 noise levels. ``drop'' columns show accuracy change from $k{=}0$ to $k{=}9$. Paired bootstrap 95\% CIs for all drops are in Appendix data.}
\label{tab:noise_per_model}
\begin{tabular}{@{}lrrrrrrrr@{}}
\toprule
& \multicolumn{4}{c}{Raw conversation} & \multicolumn{4}{c}{\ChatGPT{}-mem (NL)} \\
\cmidrule(lr){2-5}\cmidrule(lr){6-9}
Model & $k{=}0$ & $k{=}3$ & $k{=}9$ & drop & $k{=}0$ & $k{=}3$ & $k{=}9$ & drop \\
\midrule
\texttt{claude-opus-4.6}        & 60.0 & 54.0 & 55.0 & $-5.0$  & 50.5 & 47.5 & 45.0 & $-5.5$ \\
\texttt{claude-opus-4.1}        & 56.5 & 50.0 & 45.5 & $-11.0$ & 52.5 & 47.0 & 46.0 & $-6.5$ \\
\texttt{claude-sonnet-4}        & 50.5 & 45.0 & 33.0 & $-17.5$ & 47.0 & 48.5 & 48.5 & $+1.5$ \\
\texttt{gpt-5.4-mini}           & 33.5 & 27.0 & 24.5 & $-9.0$  & 38.5 & 40.0 & 36.0 & $-2.5$ \\
\texttt{gpt-5.2}                & 46.5 & 40.5 & 36.5 & $-10.0$ & 48.0 & 45.0 & 46.0 & $-2.0$ \\
\texttt{gpt-5.2-pro}            & 58.5 & 55.5 & 54.0 & $-4.5$  & 49.0 & 45.0 & 44.0 & $-5.0$ \\
\texttt{gemini-3-flash}         & 28.5 & 27.0 & 27.0 & $-1.5$  & 17.5 & 24.5 & 22.5 & $+5.0$ \\
\texttt{gemini-2.5-flash-lite}  & 31.0 & 24.5 & 19.0 & $-12.0$ & 39.5 & 42.0 & 41.5 & $+2.0$ \\
\texttt{gemini-3.1-flash-lite}  & 39.5 & 32.5 & 30.0 & $-9.5$  & 48.5 & 48.0 & 48.0 & $-0.5$ \\
\midrule
Aggregate                       & 44.9 & 39.6 & 36.1 & $-8.8$  & 43.4 & 43.1 & 41.9 & $-1.5$ \\
\bottomrule
\end{tabular}
\end{table*}

NL is more stable than raw from $k{=}0$ to $k{=}9$ on 7 of 9 models (all except \texttt{claude-opus-4.6} and \texttt{gemini-3-flash}). Paired bootstrap 95\% CIs on the NL-over-raw gap at $k{=}9$: significantly positive on 4 (\texttt{claude-sonnet-4} $+15.5$, \texttt{gpt-5.4-mini} $+11.5$, \texttt{gemini-2.5-flash-lite} $+22.5$, \texttt{gemini-3.1-flash-lite} $+18.0$pp); not significant on 3 (\texttt{claude-opus-4.1} $+0.5$, \texttt{gpt-5.2} $+9.5$, \texttt{gemini-3-flash} $-4.5$pp); negative, not significant on 1 (\texttt{claude-opus-4.6} $-10.0$pp [$-20.0,+0.5$]); significantly negative on 1 (\texttt{gpt-5.2-pro} $-10.0$pp [$-20.0,-0.5$]). Only \texttt{gpt-5.2-pro} repeats the raw-over-NL preference in the \HotpotQA evidence-surface transfer experiment; \texttt{claude-opus-4.6} does not, so the reader-subclass pattern is suggestive rather than a fixed model list (\S\ref{app:hotpot_transfer}).

\section{\HotpotQA Evidence-Surface Transfer: Per-Model Detail}
\label{app:hotpot_transfer}

Section~\ref{sec:format_transfer} reports the aggregate \HotpotQA result; Table~\ref{tab:hotpot_per_model} gives the per-model breakdown.

\begin{table}[!htbp]
\centering
\small
\setlength{\tabcolsep}{3pt}
\caption{Per-model accuracy on \HotpotQA (200 validation questions, distractor setting) across four evidence-surface renderings. Aggregate accuracy follows NL $>$ raw $>$ summary $>$ typed; the top-level NL preference transfers, while the lower-order summary/typed ranking differs from the \LongMemEval template-family comparison.}
\label{tab:hotpot_per_model}
\begin{tabular}{@{}lrrrr@{}}
\toprule
Model & Raw & NL & Sum. & Typed \\
\midrule
\texttt{claude-opus-4.6}        & 80.0 & \textbf{82.0} & 79.5 & 78.0 \\
\texttt{claude-opus-4.1}        & \textbf{86.5} & 74.5 & 69.0 & 72.0 \\
\texttt{claude-sonnet-4}        & \textbf{83.5} & 73.5 & 72.5 & 67.0 \\
\texttt{gpt-5.4-mini}           & 58.5 & 69.0 & \textbf{71.0} & 65.0 \\
\texttt{gpt-5.2}                & 74.0 & \textbf{76.5} & 72.0 & 66.5 \\
\texttt{gpt-5.2-pro}            & \textbf{77.5} & 75.0 & 74.0 & 65.5 \\
\texttt{gemini-3-flash}         & 48.0 & \textbf{50.0} & 44.0 & 39.5 \\
\texttt{gemini-2.5-flash-lite}  & 48.5 & \textbf{61.0} & 58.5 & 51.0 \\
\texttt{gemini-3.1-flash-lite}  & 61.0 & \textbf{66.5} & 64.0 & 59.5 \\
\midrule
Aggregate                       & 68.6 & \textbf{69.8} & 67.2 & 62.7 \\
\bottomrule
\end{tabular}
\end{table}

NL beats typed on all 9 models: 6 significant under paired bootstrap ($+6.5$ to $+10.5$pp), 3 within sampling noise (\texttt{claude-opus-4.6}, \texttt{claude-opus-4.1}, \texttt{gpt-5.4-mini}). NL beats summary on 8/9 models (\texttt{gpt-5.4-mini} is $-2.0$pp ns). NL beats raw on 6/9 models (3 significant: \texttt{gpt-5.4-mini} $+10.5$, \texttt{gemini-2.5-flash-lite} $+12.5$, \texttt{gemini-3.1-flash-lite} $+5.5$pp); \texttt{claude-opus-4.1} ($-12.0$pp) and \texttt{claude-sonnet-4} ($-10.0$pp) prefer raw Wikipedia paragraphs, both significantly. \texttt{gpt-5.2-pro} narrowly prefers raw ($-2.5$pp ns), matching its retrieval-noise behavior in \S\ref{app:retrieval_noise}. The \Anthropic reversals do not appear on conversational-dialogue \LongMemEval, so the reader-class hypothesis in \S\ref{sec:mechanism-hypotheses} is not fully general across domains.

\paragraph{Rendering setup.}
The summary and typed renderings are generated once per question by \texttt{claude-opus-4.6} at \texttt{temperature=0} and held fixed across evaluator models, so reader behavior is isolated from generator stochasticity. The NL surface is rule-based extraction of supporting-fact sentences (no LLM in the loop). \texttt{claude-opus-4.6} is both the generator and one of the evaluators; a generator/reader bias is possible in principle but Opus-4.6 does not systematically outperform other readers on the summary or typed conditions (Table~\ref{tab:hotpot_per_model}).

\section{Future Work}
\label{app:future}

\paragraph{End-to-end retriever integration.}
The retrieval-noise experiment (\S\ref{sec:retrieval_noise}) injects $k$ random distractors at the memory-store level. A deployed retriever produces query-correlated errors, not random ones. Running \RENDER through a dense retriever at fixed recall budgets (e.g., 1.0 down to 0.5) would test whether the evidence-surface pattern survives realistic retrieval error rather than only random noise.

\paragraph{Code and table QA.}
We show NL-over-typed ordering on conversational memory (\LongMemEval) and multi-hop Wikipedia QA (\HotpotQA). A cleaner test of the pretraining-distribution-alignment hypothesis in \S\ref{sec:mechanism-hypotheses} would run the same ladder on code- or table-grounded tasks where the typed surface dominates pretraining. If typed beats NL on those surfaces the pattern supports distribution alignment over the safety-heuristic account.

\end{document}